\documentclass[a4paper,fleqn]{cas-sc}

\usepackage[numbers]{natbib}
\usepackage{amsmath,amssymb}
\usepackage{booktabs}
\usepackage{multirow}
\usepackage{graphicx}
\usepackage{xcolor}
\usepackage{siunitx}
\usepackage{enumitem}
\usepackage{placeins}
\usepackage{capt-of}

\begin{document}
\let\WriteBookmarks\relax
\def\floatpagepagefraction{1}
\def\textpagefraction{.001}
\renewcommand{\topfraction}{0.95}
\renewcommand{\bottomfraction}{0.85}
\renewcommand{\textfraction}{0.05}
\renewcommand{\floatpagefraction}{0.95}
\setcounter{topnumber}{5}
\setcounter{bottomnumber}{3}
\setcounter{totalnumber}{8}

\shorttitle{}
\shortauthors{Huang, Luo et~al.}

\title[mode=title]{Tyan-WP: A Wind Power Foundation Model for Ultra-Short-Term Probabilistic Forecasting}

\author[a1]{Jiahui Huang}
\fnmark[$\dagger$]

\author[a1]{Ao Luo}
\fnmark[$\dagger$]

\author[a1]{Lei Liu}[orcid=0000-0002-0625-6248]
\cormark[1]
\ead{liulei13@ustc.edu.cn}

\author[a1]{Hongwei Zhao}

\author[a1]{Tengyuan Liu}

\author[a1]{Ruibo Guo}

\author[a2]{Bo Wang}

\author[a2]{Zhao Wang}

\author[a1]{Bin Li}

\cortext[1]{Corresponding author.}
\nonumnote{\textsuperscript{$\dagger$}These authors contributed equally to this work.}

\affiliation[a1]{organization={School of Information Science and Technology, University of Science and Technology of China},
            city={Hefei},
            postcode={230022},
            country={China}}

\affiliation[a2]{organization={China Electric Power Research Institute},
            city={Beijing},
            postcode={100192},
            country={China}}

\begin{abstract}
Global wind power capacity, especially in China, is booming, with new farms spanning diverse terrains and climates. The industry urgently needs accurate wind power foundation models to shorten commissioning and accelerate grid connection. This is because site-specific time series models (TSMs) are not well suited to data-scarce scenarios and generalize poorly, while generic large time series models (LTSMs) are mostly limited to univariate inputs and cannot fully exploit static site attributes or the dependencies between power and meteorological covariates, leading to insufficient accuracy. To fill this gap, we propose \textbf{Tyan-WP}, the first wind power foundation model for ultra-short-term probabilistic forecasting. Pretrained on a large-scale wind power dataset covering more than 126,000 U.S. sites over seven years, Tyan-WP further improves zero-shot forecasting through two domain-specific module designs: static site embedding using coordinate, terrain, and ecoregion metadata, and a power-aware meteorological fusion (PAMF) module that models interactions between historical power and meteorological covariates. Under a unified evaluation protocol, Tyan-WP surpasses eight site-specific supervised TSMs on 10 in-domain sites and outperforms eleven generic LTSMs on 127 in-domain sites, reducing MAE by 19.9\%, RMSE by 16.6\%, CRPS by 22.2\%, and AQL by 21.7\%, while raising $\mathrm{R^2}$ by 16.7\%. It further demonstrates strong cross-geography generalization on six real U.K. sites. These results show that the wind power foundation model can achieve accurate zero-shot forecasting without target-site training, providing a practical pathway for rapid turbine onboarding and probabilistic risk management at new wind farms.
\end{abstract}




\maketitle

\section{Introduction}\label{sec:introduction}

The global wind power industry is entering a new stage of large-scale development. According to the Global Wind Energy Council (GWEC)'s \textit{Global Wind Report 2026}, global new wind power installations reached 165 GW in 2025, with cumulative installed capacity hitting 1,299 GW \citep{gwec2026globalwind}. China's contribution is particularly notable, with data from the National Energy Administration showing that in 2025, China added 120 GW of new wind power installations \citep{nea2026windsolar}, bringing cumulative capacity to over 640 GW \citep{nea2026powerstatistics}. As high-quality wind resources in easily developable areas such as plains are gradually depleted, a large number of new wind farms are being forced to expand into regions with complex terrain and variable climatic conditions. Against this backdrop, accurate zero-shot wind power forecasting technology is crucial for shortening the trial operation period of new stations, accelerating the grid connection process, and supporting risk-aware dispatch of the power system \citep{hu2024newlybuilt,wang2022alnn}.

Early WPF relied on statistical and machine-learning methods such as support vector regression, random forests, and gradient-boosted trees \citep{soman2010review,jung2014current,liu2023deepbayesian}. Wind power forecasting is complicated by nonlinear power-curve behaviour, weather covariates, complex terrain, and ramp events \citep{wang2019powercurve,liu2024featuretemporal,bulaevskaya2015complexterrain,gallego2015ramp}. These methods remain useful in low-complexity settings, but their dependence on stationarity assumptions, manually designed features, or site-specific parameterization limits their ability to represent these interacting factors. The advent of deep learning brought recurrent architectures, convolutional and hybrid networks, and subsequently attention and Transformer-based models, which improve nonlinear temporal representation learning from sufficient historical power and meteorological observations \citep{wang2021review,wang2019deep}.

Recent time series model (TSM) architectures further advance long-sequence forecasting through several diverse modelling strategies. For example, FEDformer \citep{zhou2022fedformer} introduces frequency-enhanced decomposition for long-term temporal structure; DLinear \citep{zeng2023dlinear} revisits linear decomposition as a strong forecasting baseline; NSTransformer \citep{liu2022nstransformer} addresses distribution shift through non-stationary normalization; TimesNet \citep{wu2022timesnet} transforms temporal variation into two-dimensional representations; PatchTST \citep{nie2023patchtst} uses channel-independent patch tokens for Transformer forecasting; iTransformer \citep{liu2024itransformer} applies attention over inverted variable tokens; TimeXer \citep{wang2024timexer} strengthens Transformer forecasting with exogenous-variable modelling; and TimeMixer \citep{wang2024timemixer} performs decomposable multi-scale temporal mixing. These models learn strong target-site representations when sufficient target-site supervision is available. However, newly built or data-scarce turbine-level sites often lack adequate historical power and turbine-level meteorological observation records. Meanwhile, even with sufficient observational data, the cross-site transfer performance of site-specific TSMs is not optimistic, and it is impractical to train, validate, and maintain models separately for each wind farm site \citep{hu2024newlybuilt}.

Large time series foundation models (LTSMs) introduce a different paradigm by pretraining on broad cross-domain corpora and applying a single model to unseen series in zero-shot mode. For example, MOMENT \citep{goswami2024moment} uses masked reconstruction to learn general time series representations; TimeMoE \citep{shi2024timemoe} and Timer-series models \citep{liu2024timerbase,liu2024timerxl,liu2024timer} scale decoder or MoE-style generative time series modelling; Sundial \citep{zhang2025sundial} uses generative modelling for probabilistic time series forecasting; TiRex \citep{das2025tirex} emphasizes enhanced in-context zero-shot forecasting across long and short horizons; TimesFM \citep{das2024timesfm} adopts patched decoder-only forecasting; Chronos and Chronos 2 \citep{ansari2024chronos,ansari2025chronos2} extend language-model-style time series tokenization toward universal forecasting; and Moirai-series models \citep{woo2024moiraiunified,liu2024moiraimoe,woo2024moirai} develop universal and sparse-expert time series forecasting Transformers. LTSMs provide a possible zero-shot route, but they are designed for generic time series, and wind power accounts for only a small fraction of their broad pretraining corpora. As a result, their objectives do not naturally favour wind-power data distributions. Moreover, most public LTSMs expose only single-series sequence (S3)-style interfaces, namely univariate input-only formats, and therefore cannot exploit the dependencies between historical power and meteorological covariates \citep{goswami2024moment,shi2024timemoe,liu2024timerbase,das2024timesfm}. More generally, generic LTSMs do not explicitly encode domain-specific structural priors such as meteorological covariates and complex land-surface representations, which are important for wind power forecasting \citep{wang2019powercurve,bulaevskaya2015complexterrain}. Table~\ref{tab:wpf_temporal_summary} summarises representative time series modelling methods considered in this study.

\begin{table}[!t]
\caption{The summary of selected WPF research based on time series modeling.}\label{tab:wpf_temporal_summary}
\centering
\scriptsize
\setlength{\tabcolsep}{2.2pt}
\renewcommand{\arraystretch}{0.90}
\begin{tabular*}{\textwidth}{@{\extracolsep{\fill}}p{0.21\textwidth}p{0.17\textwidth}p{0.41\textwidth}p{0.12\textwidth}@{}}
\toprule
Classification & Method & Authors \& Organization & Release time \\
\midrule
\multirow{8}{0.21\textwidth}{Deep-learning-based time series models}
& FEDformer \citep{zhou2022fedformer} & Zhou et al. (Alibaba Group) & 2022-01 \\
& DLinear \citep{zeng2023dlinear} & Zeng, Chen et al. (The Chinese University of Hong Kong; IDEA) & 2022-05 \\
& NSTransformer \citep{liu2022nstransformer} & Liu, Wu et al. (Tsinghua University) & 2022-05 \\
& TimesNet \citep{wu2022timesnet} & Wu et al. (Tsinghua University) & 2022-10 \\
& PatchTST \citep{nie2023patchtst} & Nie et al. (IBM Research) & 2022-11 \\
& iTransformer \citep{liu2024itransformer} & Liu et al. (Tsinghua University) & 2023-10 \\
& TimeXer \citep{wang2024timexer} & Wang et al. (Tsinghua University) & 2024-02 \\
& TimeMixer \citep{wang2024timemixer} & Wang et al. (Tsinghua University) & 2024-05 \\
\midrule
\multirow{17}{0.21\textwidth}{Generic large time series models}
& Timer \citep{liu2024timerbase} & Liu et al. (Tsinghua University) & 2024-02 \\
& MOMENT \citep{goswami2024moment} & Goswami et al. (Carnegie Mellon University) & 2024-02 \\
& Chronos \citep{ansari2024chronos} & Ansari et al. (Amazon Web Services) & 2024-03 \\
& Moirai 1.0 \citep{woo2024moiraiunified} & Woo, Liu et al. (Salesforce AI Research) & 2024-03 \\
& Moirai 1.1 \citep{woo2024moiraiunified} & Woo, Liu et al. (Salesforce AI Research) & 2024-06 \\
& TimesFM 1.0 \citep{das2024timesfm} & Das et al. (Google Research) & 2024-08 \\
& TimeMoE \citep{shi2024timemoe} & Shi, Wang, Nie et al. (Princeton University; Squirrel AI Learning; Griffith University) & 2024-09 \\
& Timer-XL \citep{liu2024timerxl} & Liu et al. (Tsinghua University) & 2024-10 \\
& Moirai-MoE \citep{liu2024moiraimoe} & Liu, Liu, Woo et al. (Salesforce AI Research) & 2024-10 \\
& Chronos-Bolt \citep{ansari2024chronos} & Ansari et al. (Amazon Web Services) & 2024-11 \\
& TimesFM 2.0 \citep{das2024timesfm} & Das et al. (Google Research) & 2024-12 \\
& Sundial \citep{zhang2025sundial} & Liu et al. (Tsinghua University) & 2025-02 \\
& TiRex \citep{das2025tirex} & Auer et al. (Johannes Kepler University Linz) & 2025-05 \\
& TimesFM 2.5 \citep{das2024timesfm} & Das et al. (Google Research) & 2025-09 \\
& Chronos 2 \citep{ansari2025chronos2} & Ansari, Shchur, K{\"u}ken et al. (Amazon Web Services) & 2025-10 \\
& Moirai 2.0 \citep{woo2024moirai} & Liu, Aksu, Liu, Liu et al. (Salesforce AI Research) & 2025-11 \\
& Timer-S1 \citep{liu2024timer} & Liu, Su, Wang, Zhang et al. (Tsinghua University) & 2026-03 \\
\midrule
Wind power foundation models
& Tyan-WP (Ours) & Huang, Luo et al. (University of Science and Technology of China) & This work \\
\bottomrule
\end{tabular*}
\end{table}

The preceding comparison suggests that two types of domain-specific structural information help to perform accurate zero-shot WPF for all turbine sites of newly built wind farms: (1) static site-level information, used to represent geographical location, terrain class, and ecoregion context, which affect local wind regimes, turbine responses, and cross-site transfer; and (2) dynamic variable-level information, used to model the dependencies between historical power and meteorological covariates, including nonlinear wind-speed-to-power conversion, wind-direction periodicity, air-density effects, ramp events, and power inertia.

These requirements motivate a domain-specific foundation model rather than a direct reuse of generic LTSMs. Domain-specific foundation models provide a practical route because they combine transferable pretraining with task-specific data, variables, and architectural priors. Recent examples include PriceFM for electricity price forecasting \citep{yu2025pricefm}, Kronos for financial-market forecasting \citep{shi2025kronos}, and MIRA for medical time series modelling \citep{li2025mira}. These studies indicate that general-purpose temporal pretraining can be substantially strengthened when the model is redesigned around the structure of a target domain. Nevertheless, the field of wind power forecasting still lacks a dedicated pre-trained model capable of probabilistic forecasting by jointly utilizing historical power data, meteorological covariates, and site metadata, enabling zero-shot deployment at newly built or data-scarce wind turbine sites with varying topography and meteorological conditions.

To fill this gap, this paper proposes \textbf{Tyan-WP}, a wind power foundation model that is pretrained on a large-scale wind power dataset containing over 126,000 wind turbine sites, designed for zero-shot probabilistic forecasting at new or data-scarce turbine-level sites. Tyan-WP targets newly built wind turbine sites where long historical power records are insufficient for reliable supervised training, while recent local power and meteorological measurements are available at inference time. Methodologically, Tyan-WP represents historical power and meteorological histories through dual-branch patch embeddings, derives time-level calendar embeddings from timestamps, and constructs site-level geography-ecology embeddings from static metadata. It then uses a power-aware meteorological fusion (PAMF) module that models dependencies between recent power dynamics and meteorological covariates, processes the fused sequence with a sparse expert encoder, and directly outputs multi-horizon quantile forecasts through a horizon-quantile forecasting head.

The main contributions are as follows:
\begin{enumerate}[label=(\arabic*)]
\item \textbf{Wind power foundation model.} We propose Tyan-WP, the first wind power foundation model for ultra-short-term probabilistic forecasting to the best of our knowledge, supporting zero-shot deployment at newly built or data-scarce wind turbine sites.
\item \textbf{Domain-specific module design.} In addition to pretraining on a large-scale wind power dataset, Tyan-WP further improves zero-shot forecasting through two domain-specific module designs: static site embedding using latitude, longitude, terrain, and ecoregion metadata, and a power-aware meteorological fusion (PAMF) module that models the dependencies between historical power and meteorological covariates.
\item \textbf{Optimal prediction performance.} Under in-domain and out-of-domain zero-shot evaluation protocols, Tyan-WP achieves the best overall performance among the evaluated site-specific time series models (TSMs) and generic large time series models (LTSMs), improving deterministic accuracy and probabilistic forecast quality across all benchmark settings.
\item \textbf{Practical deployment value.} With the advantage of not requiring training at the target site, Tyan-WP directly shortens the model deployment and trial operation cycles for new wind farms. It not only enables rapid commissioning of wind turbines and accelerates grid integration, but also provides a technical foundation for probabilistic risk management in power systems.
\end{enumerate}


\section{Preliminary}\label{sec:preliminary}

\subsection{Problem formulation}\label{sec:problem}

This paper defines ultra-short-term wind power forecasting as a multivariate time series probabilistic forecasting problem. Given a wind turbine site, the forecasting task is defined over a discrete time axis with 15-minute resolution. Let $\mathbf{P}_{t-L+1:t}\in\mathbb{R}^{L\times 1}$ denote the historical power series, $\mathbf{M}_{t-L+1:t}\in\mathbb{R}^{L\times C_m}$ the historical meteorological series with $C_m=6$ channels (corresponding to wind speed, wind direction sine, wind direction cosine, air density, temperature, and surface air pressure, respectively), $\boldsymbol{\tau}_t$ the timestamp information, and $S_{site}$ the static site metadata (including longitude, latitude, terrain type, ecoregion). The objective is to predict the conditional quantile function of the future power:
\begin{equation}\label{eq:task}
  \hat{\mathbf{Y}} = f_\theta\!\bigl(\mathbf{P}_{t-L+1:t},\;\mathbf{M}_{t-L+1:t},\;\boldsymbol{\tau}_t,\;S_{site}\bigr) \;\in\;\mathbb{R}^{H\times|\mathcal{Q}|},
\end{equation}
where $L=64$ is the input window length (16\,h), $H=16$ is the forecast horizon (4\,h), and $\mathcal{Q}=\{0.1,0.2,\ldots,0.9\}$ is the set of nine target quantile levels.

The point forecast is taken as the median quantile $q_{0.5}$. Hard constraints enforced across all models and experiments are: (i) no future meteorological variables are used as input; (ii) all evaluation metrics are computed in the original MW power scale; (iii) the prediction target is turbine-level or turbine-level-site power, not aggregated wind-farm power lacking turbine-level sensing.

\subsection{Data and preprocessing}\label{sec:data_preprocessing}

\subsubsection{Dataset description}\label{sec:dataset_description}

\begin{figure}
  \centering
  \includegraphics[width=\textwidth]{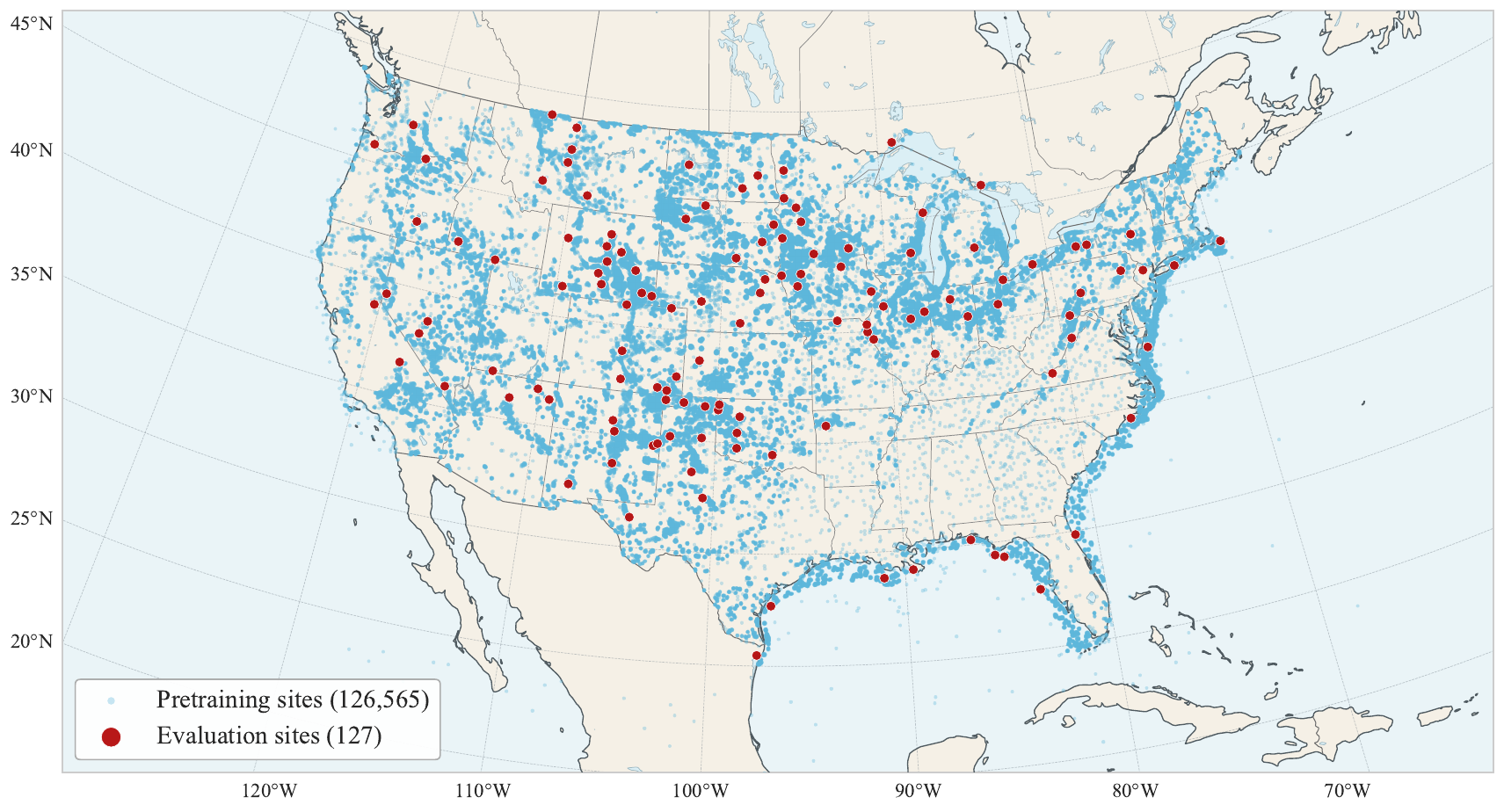}
  \caption{Geographical distribution of all sites in the WTK dataset, covering the continental United States, categorized by pretraining and evaluation pools. Unseen sites within the domain (used for evaluation) are selected randomly and represent various ecological regions, including inland and coastal areas.}
  \label{fig:wtk_conus_site_distribution}
\end{figure}

\begin{figure}
  \centering
  \includegraphics[width=\textwidth]{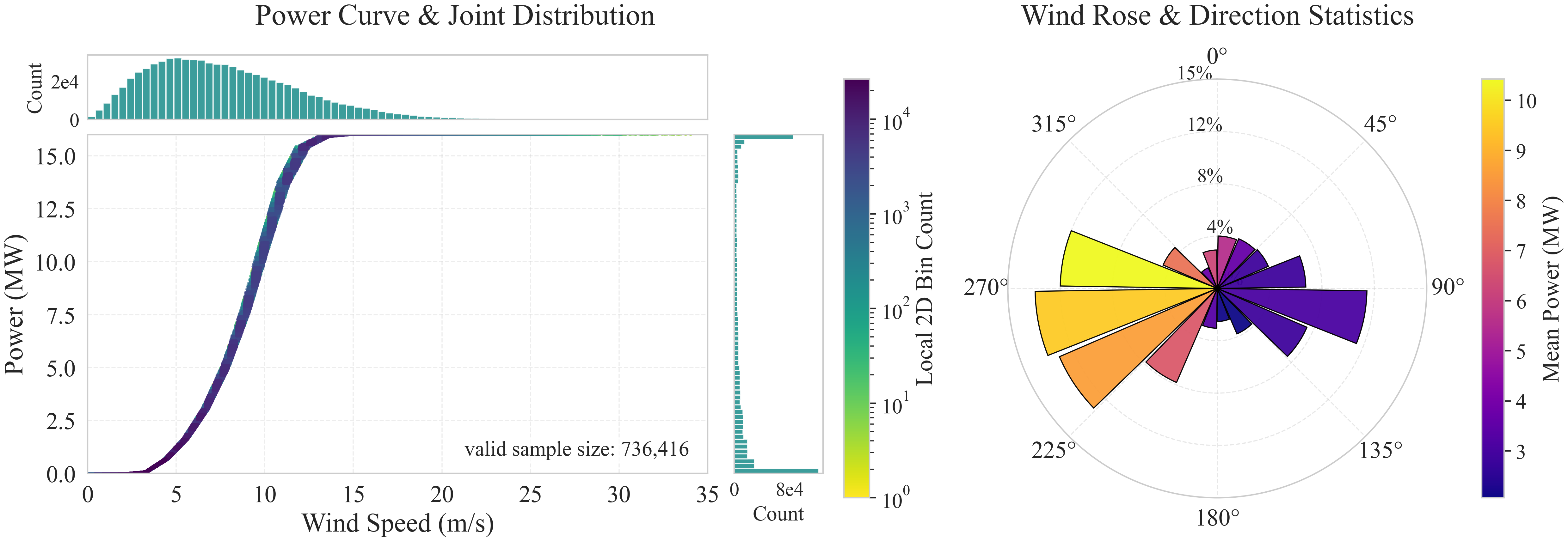}
  \caption{Visualization of power curve and wind rose at a representative site in the WTK dataset. $\mathbf{Left}$: joint distribution of wind speed and turbine-level power at site 9094, with marginal histograms and a two-dimensional bin-count colour scale summarising 736,416 valid samples. $\mathbf{Right}$: wind rose summarising the directional frequency and wind-speed distribution at the same site.}
  \label{fig:site_9094_joint_distribution_and_wind_rose}
\end{figure}

The NREL WIND Toolkit (WTK) \citep{draxl2015wind} provides simulated wind power and meteorological variables for \num{126692} sites across the continental United States at 5-minute resolution over 2007--2013. Each site records power output alongside five meteorological variables: wind speed, wind direction, air density, temperature, and surface air pressure.

Figure~\ref{fig:wtk_conus_site_distribution} shows that the WTK sites cover a broad geographical range, including the Great Plains, coastal corridors, mountainous regions, and offshore-influenced areas. This large spatial span exposes the model to diverse wind climates, land-surface conditions, and turbine operating regimes during pretraining. At the same time, the held-out in-domain evaluation sites remain geographically dispersed rather than concentrated in a narrow subregion, which makes the subsequent zero-shot tests more representative of practical deployment on unseen turbines.

Figure~\ref{fig:site_9094_joint_distribution_and_wind_rose} further illustrates the statistical structure of a representative WTK site. The left panel shows the characteristic nonlinear wind-speed-to-power relationship, including a low-power regime at weak wind speeds, a steep transition region, and a high-power saturation regime near rated output. The spread of observations in the intermediate wind-speed range also indicates that wind speed alone is insufficient to determine power uniquely, motivating the inclusion of additional meteorological variables and temporal context. The right panel shows a directional concentration of wind occurrence together with direction-dependent wind-speed frequencies, suggesting that local wind regimes are anisotropic rather than isotropic. These properties support the design choice of learning wind-specific representations from both multivariate meteorological histories and site descriptors.

\subsubsection{Data preprocessing}\label{sec:preprocessing}

Considering the time resolution required for practical operational ultra-short-term wind power forecasting, the original 5-minute WTK data are averaged and resampled to a uniform 15-minute resolution. Wind direction is decomposed into sine and cosine components to avoid the $0^\circ$/$360^\circ$ discontinuity. Meteorological variables are standardised via global z-score normalisation using pretraining-set statistics. Power values are normalised by the rated capacity (16.0\,MW). Obvious high-wind-speed low-power anomalies are corrected by rule-based filtering. Non-finite and unphysical values are removed, and sliding windows containing any NaN entries are excluded from both training and evaluation.

\section{Methodology}\label{sec:method}

\begin{figure}
  \centering
  \includegraphics[width=\textwidth]{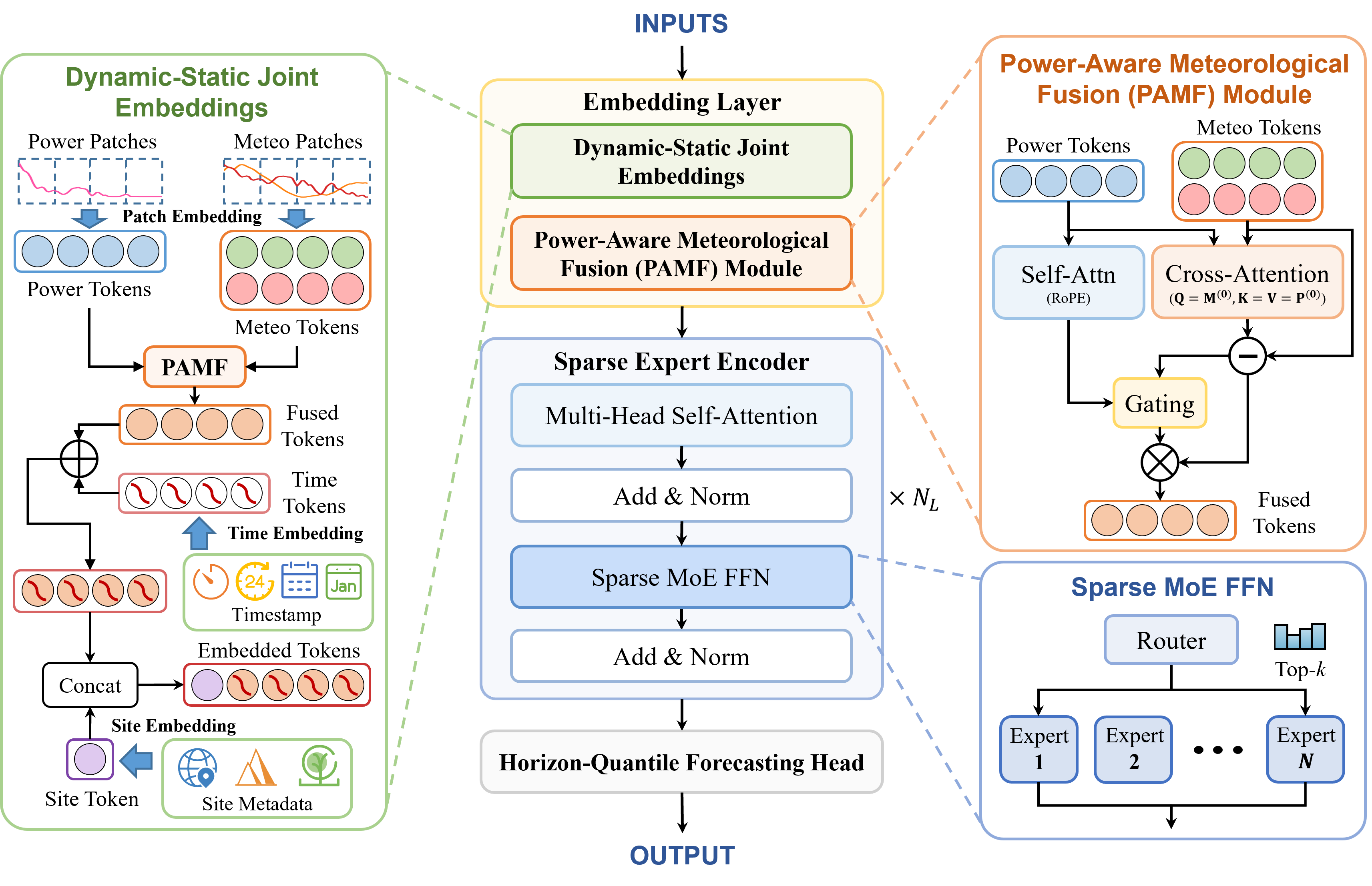}
  \caption{The overall framework of Tyan-WP. The model takes historical power sequences, historical meteorological sequences, timestamp information, and static site metadata as inputs, without using future meteorological variables. Dynamic sequences are mapped into dual-branch patch embeddings, while timestamps and static metadata are mapped into time-level calendar embeddings and site-level geography-ecology embeddings, respectively. The power-aware meteorological fusion (PAMF) module models interactions between recent power dynamics and historical meteorological covariates to generate fused representations. The sparse expert encoder then combines relative temporal phase modelling with sparse expert routing to capture heterogeneous wind-regime patterns. Finally, the horizon-quantile forecasting head produces multi-horizon quantile forecasts for ultra-short-term probabilistic wind power forecasting.}
  \label{fig:overall_framework}
\end{figure}

Tyan-WP is designed for zero-shot ultra-short-term wind power forecasting at newly built wind turbine sites lacking historical observation data. For each forecasting window, the model takes four types of inputs: the historical power sequence, the historical meteorological sequence, timestamp information, and static site metadata, including latitude, longitude, terrain type, and ecoregion. Future meteorological variables are not used in either training or evaluation. Tyan-WP first maps dynamic sequences, timestamps, and static descriptors into dual-branch patch embeddings, time-level calendar embeddings, and site-level geography-ecology embeddings, respectively. It then applies a power-aware meteorological fusion (PAMF) module to model dependencies between recent power dynamics and historical meteorological covariates, routes the resulting sequence through a sparse expert encoder, and directly outputs multi-horizon quantile forecasts through the horizon-quantile forecasting head. Figure~\ref{fig:overall_framework} shows the overall framework of Tyan-WP.

\subsection{Dynamic-static joint embeddings}\label{sec:dynamic_static_embeddings}

Tyan-WP uses dynamic-static joint embeddings to map heterogeneous forecasting inputs into a common token space before sequence modelling. The dual-branch patch embedding module represents historical power and meteorological sequences while preserving their different physical roles. The time-level calendar embedding module provides periodic timestamp information, and the site-level geography-ecology embedding module introduces static site priors related to location, terrain, and ecoregion.

\subsubsection{Dual-branch patch embedding}\label{sec:patch_embedding}

Let $\mathbf{P}\in\mathbb{R}^{L\times 1}$ denote the historical power sequence and $\mathbf{M}\in\mathbb{R}^{L\times C_m}$ denote the historical meteorological sequence. Tyan-WP uses a dual-branch tokenizer rather than early channel concatenation, so that turbine response inertia and exogenous meteorological forcing are preserved as distinct streams before fusion. Each branch is segmented into patches of length $p$ with stride $s$, producing
\begin{equation}
  T_p=\left\lfloor\frac{L-p}{s}\right\rfloor+1
\end{equation}
patches. For patch index $j\in\{1,\dots,T_p\}$, the embedding step is
\begin{align}
  \mathbf{p}_j &= \mathrm{LN}\!\left(\mathbf{W}_{p}\,\mathrm{vec}\!\left(\mathbf{P}_{(j)}\right)+\mathbf{b}_{p}\right), \\
  \mathbf{m}_j &= \mathrm{LN}\!\left(\mathbf{W}_{m}\,\mathrm{vec}\!\left(\mathbf{M}_{(j)}\right)+\mathbf{b}_{m}\right),
\end{align}
where $\mathbf{P}_{(j)}\in\mathbb{R}^{p\times 1}$ and $\mathbf{M}_{(j)}\in\mathbb{R}^{p\times C_m}$ are the $j$-th sliced patches, $\mathrm{vec}(\cdot)$ denotes vectorization that flattens each patch into a one-dimensional token in temporal order, $\mathrm{LN}(\cdot)$ denotes layer normalization, $\mathbf{p}_j\in\mathbb{R}^{d}$ denotes the $j$-th power-patch token, $\mathbf{m}_j\in\mathbb{R}^{d}$ denotes the corresponding meteorological-patch token, and $d$ denotes the token embedding dimension. Stacking all tokens yields
\begin{equation}
  \mathbf{P}^{(0)}=[\mathbf{p}_1,\dots,\mathbf{p}_{T_p}]^\top,\qquad
  \mathbf{M}^{(0)}=[\mathbf{m}_1,\dots,\mathbf{m}_{T_p}]^\top \in \mathbb{R}^{T_p\times d}.
\end{equation}
No learned absolute positional embedding is added at this stage. Instead, temporal information is introduced through time-level calendar embeddings and RoPE, which avoids binding the input representation to a fixed absolute index space.

\subsubsection{Time-level calendar embedding}\label{sec:calendar_embedding}

For each patch index $j$, the calendar embedding module uses the ending timestamp $t^{\mathrm{end}}_j$ of the corresponding patch rather than the starting timestamp, so that the temporal tag aligns with the latest observation contained in that patch. Let $\mu_j$, $h_j$, $d_j$, and $m_j$ denote the minute, hour, day-of-month, and month extracted from $t^{\mathrm{end}}_j$, and let $D_j$ denote the number of days in month $m_j$. The calendar descriptor is
\begin{equation}
  \mathbf{c}_j=
  \begin{bmatrix}
    \sin(2\pi \mu_j / 60), & \cos(2\pi \mu_j / 60), &
    \sin(2\pi h_j / 24), & \cos(2\pi h_j / 24), \\
    \sin(2\pi (d_j-1) / D_j), & \cos(2\pi (d_j-1) / D_j), &
    \sin(2\pi (m_j-1) / 12), & \cos(2\pi (m_j-1) / 12)
  \end{bmatrix}.
\end{equation}
Here, $\mathbf{c}_j$ is a continuous calendar descriptor rather than a discrete time label. Accordingly, $\mathrm{MLP}_{\mathrm{time}}(\cdot)$ is implemented as a two-layer multilayer perceptron with a nonlinear activation between the two linear projections:
\begin{equation}
  \mathrm{MLP}_{\mathrm{time}}(\mathbf{c}_j)=
  \mathbf{W}^{(2)}_{t}\,\phi\!\left(\mathbf{W}^{(1)}_{t}\mathbf{c}_j+\mathbf{b}^{(1)}_{t}\right)+\mathbf{b}^{(2)}_{t},
\end{equation}
where $\phi(\cdot)$ denotes the activation function. The corresponding time-level calendar embedding is
\begin{equation}
  \mathbf{e}^{\mathrm{time}}_j=\mathrm{LN}\!\left(\mathrm{MLP}_{\mathrm{time}}(\mathbf{c}_j)\right).
\end{equation}
This embedding provides explicit periodic cues for sub-daily and seasonal variability without introducing hand-crafted forecast rules.

\subsubsection{Site-level geography-ecology embedding}\label{sec:site_embedding}

Let $\varphi$ and $\lambda$ denote latitude and longitude, and let $c_{\mathrm{terr}}$ and $c_{\mathrm{eco}}$ denote terrain and ecoregion categories. Tyan-WP maps static site metadata to a single geography-ecology embedding through
\begin{align}
  \mathbf{g}^{\mathrm{geo}} &= \left[\sin\varphi,\cos\varphi,\sin\lambda,\cos\lambda\right], \\
  \mathbf{e}^{\mathrm{terr}} &= \mathrm{Emb}_{\mathrm{terr}}(c_{\mathrm{terr}}), \\
  \mathbf{e}^{\mathrm{eco}} &= \mathrm{Emb}_{\mathrm{eco}}(c_{\mathrm{eco}}), \\
  \mathbf{e}^{\mathrm{site}} &= \mathrm{LN}\!\Bigl(\mathbf{W}_{s}\bigl[
  \mathbf{W}_{g}\mathbf{g}^{\mathrm{geo}};\,
  \mathbf{e}^{\mathrm{terr}};\,
  \mathbf{e}^{\mathrm{eco}}
  \bigr]+\mathbf{b}_{s}\Bigr).
\end{align}
Unlike $\mathrm{MLP}_{\mathrm{time}}(\cdot)$, which learns a nonlinear projection from continuous temporal descriptors, $\mathrm{Emb}_{\mathrm{terr}}(\cdot)$ and $\mathrm{Emb}_{\mathrm{eco}}(\cdot)$ are learnable embedding lookups for discrete site categories. If $\mathbf{E}_{\mathrm{terr}}$ and $\mathbf{E}_{\mathrm{eco}}$ denote the terrain and ecoregion embedding matrices, then
\begin{align}
  \mathbf{e}^{\mathrm{terr} }&= \mathbf{E}_{\mathrm{terr}}[c_{\mathrm{terr}}], \\
  \mathbf{e}^{\mathrm{eco}} &= \mathbf{E}_{\mathrm{eco}}[c_{\mathrm{eco}}],
\end{align}
where the bracket notation denotes row selection by category index. Therefore, the calendar embedding module and the site-category embedding lookups serve different roles and are implemented differently, even though they are later used in the same token space.

\subsection{Power-aware meteorological fusion (PAMF) module}\label{sec:meteofusion}

Simple concatenation treats all meteorological information as equally useful, even though part of it overlaps with information already reflected in the recent power trajectory. To make meteorological features more informative for zero-shot forecasting, the power-aware meteorological fusion (PAMF) module first strengthens short-term power dynamics and then uses power-conditioned cross-attention to derive complementary meteorological features. Using rotary self-attention and cross-attention in the standard query--key--value form \citep{vaswani2017attention}, the module computes
\begin{align}
  \widetilde{\mathbf{P}} &= \mathrm{LN}\!\left(\mathbf{P}^{(0)} + \mathrm{SelfAttn}\!\left(\mathbf{P}^{(0)}\right)\right), \\
  \mathbf{M}^{\mathrm{cond}} &= \mathrm{LN}\!\left(\mathrm{CrossAttn}\!\left(\mathbf{Q}=\mathbf{M}^{(0)},\mathbf{K}=\mathbf{P}^{(0)},\mathbf{V}=\mathbf{P}^{(0)}\right)\right), \\
  \mathbf{M}^{\mathrm{comp}} &= \mathbf{M}^{(0)} - \mathbf{M}^{\mathrm{cond}}.
\end{align}
Here, $\widetilde{\mathbf{P}}$ reinforces short-term power dynamics, while $\mathbf{M}^{\mathrm{cond}}$ denotes the meteorological component aligned with the current power state. The complementary component $\mathbf{M}^{\mathrm{comp}}$ captures meteorological information that is not already represented by the power branch. The power-aware meteorological fusion (PAMF) module then uses a gate to integrate these two sources:
\begin{align}
  \mathbf{G} &= \sigma\!\left(\mathbf{W}_{g}\left[\widetilde{\mathbf{P}};\mathbf{M}^{\mathrm{comp}}\right]+\mathbf{b}_{g}\right), \\
  \mathbf{F} &= \mathrm{LN}\!\left(\mathbf{G}\odot\widetilde{\mathbf{P}} + \left(1-\mathbf{G}\right)\odot\mathbf{M}^{\mathrm{comp}}\right),
\end{align}
where $\sigma(\cdot)$ denotes the sigmoid activation, $\odot$ denotes the Hadamard element-wise product, $\mathbf{G}\in\mathbb{R}^{T_p\times d}$ is an element-wise gate, and $\mathbf{F}\in\mathbb{R}^{T_p\times d}$ is the fused patch sequence. Writing its rows explicitly,
\begin{equation}
  \mathbf{F}=[\mathbf{f}_1,\dots,\mathbf{f}_{T_p}]^\top,
\end{equation}
where $\mathbf{f}_j\in\mathbb{R}^{d}$ denotes the fused representation of the $j$-th patch. This integration is representation-level rather than causal identification, but it explicitly encourages the model to emphasize meteorological changes that complement recent power dynamics and are informative for imminent ramps.

\subsection{Sparse expert encoder}\label{sec:encoder}

Before entering the sparse expert encoder, each fused patch representation is augmented by its time-level calendar embedding, and the site-level geography-ecology embedding is prepended as a site token:
\begin{align}
  \bar{\mathbf{f}}_j &= \mathbf{f}_j+\mathbf{e}^{\mathrm{time}}_j,  \\
  \mathbf{H}^{(0)} &= \left[\mathbf{e}^{\mathrm{site}};\bar{\mathbf{f}}_1;\dots;\bar{\mathbf{f}}_{T_p}\right]
  \in\mathbb{R}^{(T_p+1)\times d}.
\end{align}
This construction allows site priors to be broadcast to all patch tokens through subsequent self-attention, which is useful because cut-in behaviour, wake exposure, and regime frequency vary systematically with geography and land surface context.

The sparse expert encoder applies $N_L$ stacked sparse layers to $\mathbf{H}^{(0)}$. In layer $\ell$, RoPE self-attention \citep{su2021roformer} first models interactions among the site token and all patch tokens:
\begin{equation}
  \mathbf{A}^{(\ell)}=
  \mathrm{LN}^{(\ell)}\!\left(
  \mathbf{H}^{(\ell-1)} +
  \mathrm{SelfAttn}^{(\ell)}\!\left(\mathbf{H}^{(\ell-1)}\right)
  \right).
\end{equation}
RoPE preserves relative phase information along the patch sequence without relying on learned absolute position vectors, which is well aligned with short-horizon temporal dependency modelling. Each token $\mathbf{a}^{(\ell)}_j$ is then routed through a sparse Mixture-of-Experts (MoE) layer \citep{shazeer2017outrageously,fedus2022switch}:
\begin{align}
  \boldsymbol{\pi}^{(\ell)}_j &= \mathrm{softmax}\!\left(\mathbf{W}^{(\ell)}_{r}\mathbf{a}^{(\ell)}_j\right), \\
  \mathcal{K}^{(\ell)}_j &= \mathrm{TopK}\!\left(\boldsymbol{\pi}^{(\ell)}_j,k\right), \\
  \mathbf{u}^{(\ell)}_j &= \sum_{e\in\mathcal{K}^{(\ell)}_j}\tilde{\pi}^{(\ell)}_{j,e}\,
  \mathrm{FFN}^{(\ell)}_{e}\!\left(\mathbf{a}^{(\ell)}_j\right), \\
  \mathbf{h}^{(\ell)}_j &= \mathrm{LN}^{(\ell)}\!\left(\mathbf{a}^{(\ell)}_j + \mathbf{u}^{(\ell)}_j\right),
\end{align}
where $\tilde{\pi}^{(\ell)}_{j,e}$ denotes the normalized routing weight over the selected top-$k$ experts. Each expert is a two-layer feed-forward network with GELU activation and dropout. All tokens share the same router, so expert selection depends jointly on recent dynamics and site context. This sparse mechanism is intended to let different experts specialize to wind regimes such as persistent plateaus, sharp ramps, offshore-like flows, or complex-terrain responses without paying dense-compute cost for every token.

\subsection{Horizon-quantile forecasting head}\label{sec:head}

Tyan-WP uses a direct multi-horizon quantile head instead of autoregressive decoding. After the final encoder layer, only the patch tokens are passed to the forecasting head, while the site token has already influenced them through self-attention. Let
\begin{equation}
  \mathbf{u}=
  \left[\mathbf{h}^{(N_L)}_1;\dots;\mathbf{h}^{(N_L)}_{T_p}\right]
  \in\mathbb{R}^{T_p d}
\end{equation}
denote the flattened patch representation. The head predicts all horizons and quantiles in a single pass:
\begin{equation}
  \hat{\mathbf{Y}}=
  \mathrm{reshape}\!\left(\mathrm{MLP}_{\mathrm{head}}(\mathbf{u})\right)
  \in\mathbb{R}^{H\times |\mathcal{Q}|}.
\end{equation}
The entry $\hat{y}_{h,q}$ denotes the predicted $q$-th quantile at horizon $h$. This design keeps inference efficient and allows every forecast horizon to condition on the full encoded history instead of only on earlier generated outputs.

\subsection{Optimization objective}\label{sec:objective}

Training combines multi-horizon quantile regression \citep{koenker1978regression} with an auxiliary load-balancing term for sparse expert routing. For target $y_{b,h}$ and predicted quantile $\hat{y}_{b,h,q}$, the pinball loss is
\begin{equation}
  \ell_q(y_{b,h},\hat{y}_{b,h,q})=
  \max\!\left(q\left(y_{b,h}-\hat{y}_{b,h,q}\right),\left(q-1\right)\left(y_{b,h}-\hat{y}_{b,h,q}\right)\right).
\end{equation}
The quantile-regression objective over batch size $B$, horizon length $H$, and quantile set $\mathcal{Q}$ is
\begin{equation}
  \mathcal{L}_{\mathrm{quant}}=
  \frac{1}{B\,H\,|\mathcal{Q}|}
  \sum_{b=1}^{B}\sum_{h=1}^{H}\sum_{q\in\mathcal{Q}}
  \ell_q(y_{b,h},\hat{y}_{b,h,q}).
\end{equation}
For the MoE router in layer $\ell$, let
\begin{equation}
  \bar{\boldsymbol{\pi}}^{(\ell)}=
  \frac{1}{B\,(T_p+1)}
  \sum_{b=1}^{B}\sum_{j=1}^{T_p+1}\boldsymbol{\pi}^{(\ell)}_{b,j}
\end{equation}
be the mean routing distribution across all tokens, and let $\mathbf{u}_{E}$ be the uniform distribution over the $E$ experts. The auxiliary term is
\begin{equation}
  \mathcal{L}_{\mathrm{aux}}=
  \sum_{\ell=1}^{N_L}
  D_{\mathrm{KL}}\!\left(\mathbf{u}_{E}\,\|\,\bar{\boldsymbol{\pi}}^{(\ell)}\right).
\end{equation}
The final objective is
\begin{equation}
  \mathcal{L}=\mathcal{L}_{\mathrm{quant}}+\lambda\,\mathcal{L}_{\mathrm{aux}},
\end{equation}
where $\lambda$ controls the strength of the routing regularizer. This objective encourages accurate quantile forecasts while reducing expert collapse and improving regime specialization.

\section{Experimental design}\label{sec:experiment}

\subsection{Model configuration and training details}\label{sec:config_training}

All reported Tyan-WP results use the same 35.53\,M-parameter configuration. Table~\ref{tab:config} summarises the architectural and optimization-related hyperparameters used in the main experiments.

\begin{table}[!t]
\caption{Model configuration of Tyan-WP used in the main experiments. EPA: United States Environmental Protection Agency.}\label{tab:config}
\centering
\small
\begin{tabular*}{\textwidth}{@{\extracolsep{\fill}}p{0.26\textwidth}p{0.50\textwidth}p{0.16\textwidth}@{}}
\toprule
Parameter & Meaning & Value \\
\midrule
\texttt{d\_model} & Hidden dimension of patch and token representations & 256 \\
\texttt{n\_heads} & Number of attention heads & 8 \\
\texttt{d\_ff} & Hidden dimension of each expert feed-forward network & 1024 \\
\texttt{patch\_size} & Number of time steps per input patch & 8 \\
\texttt{stride} & Sliding stride between adjacent patches & 8 \\
\texttt{seq\_in} & Length of the historical input window & 64 \\
\texttt{pred\_len} & Length of the forecast horizon & 16 \\
\texttt{n\_met\_vars} & Number of historical meteorological variables & 6 \\
\texttt{n\_layers} & Number of stacked sparse expert encoder layers & 10 \\
\texttt{n\_experts} & Number of experts in each MoE layer & 6 \\
\texttt{top\_k} & Number of activated experts per token & 2 \\
\texttt{n\_terrain} & Number of terrain categories & 6 \\
\texttt{n\_eco\_l1} & Number of EPA Level I ecoregion categories & 11 \\
\texttt{n\_quantiles} & Number of predicted quantile levels & 9 \\
\texttt{dropout} & Dropout rate & 0.1 \\
\texttt{time\_mlp\_hidden} & Hidden dimension of the calendar-embedding MLP & 128 \\
\texttt{rope\_base} & Base frequency used by RoPE & 10000.0 \\
\texttt{aux\_weight} & Weight of the MoE load-balancing loss & 0.03 \\
\texttt{Total params} & Total number of trainable parameters & 35.53\,M \\
\texttt{MoE-active params} & Parameters activated under sparse expert routing & 13.17\,M \\
\texttt{Inference-active params} & Parameters traversed in one inference pass & 14.51\,M \\
\bottomrule
\end{tabular*}
\end{table}

Training uses AdamW with learning rate $5\times10^{-5}$, exponential decay rate $\beta_1=0.90$, $\beta_2=0.95$, weight decay 0.02, and linear warmup over 3\% of the total optimization steps. The batch size is 26,624, gradient clipping is set to 1.0, and the model is trained for 20 epochs. The implementation uses PyTorch~2.8.0 with CUDA~12.8. Experiments were conducted on a server with two AMD EPYC 9B14 96-core CPUs, 1.5\,TiB RAM, and eight NVIDIA RTX PRO 6000 Blackwell GPUs (96\,GB each). Public LTSM inference used a separate Python environment.

\subsection{Baseline models}\label{sec:baselines}

This paper compares Tyan-WP against two representative baseline families. The first family is site-specific time series models (TSM), which are trained in a fully supervised manner on each target wind site. The second family is large time series models (LTSM), which are pretrained on large-scale cross-domain corpora and evaluated in zero-shot mode without target-site fine-tuning.

\subsubsection{Time series models (TSM)}\label{sec:tsm_settings}

Eight representative TSM architectures are evaluated as site-specific supervised baselines on the WTK 10-site subset. These models span the major paradigms in modern time series forecasting: linear decomposition (DLinear \citep{zeng2023dlinear}), frequency-domain attention (FEDformer \citep{zhou2022fedformer}), non-stationary normalisation (NSTransformer \citep{liu2022nstransformer}), temporal 2D-variation modelling (TimesNet \citep{wu2022timesnet}), channel-independent patching (PatchTST \citep{nie2023patchtst}), inverted-dimension attention (iTransformer \citep{liu2024itransformer}), exogenous-variable Transformers (TimeXer \citep{wang2024timexer}), and multi-scale mixing (TimeMixer \citep{wang2024timemixer}). Each model is trained independently per site with five random seeds and averaged. Training uses AdamW (lr $10^{-4}$, weight decay $10^{-2}$), batch size 2048, max 200 epochs, and early stopping patience 20. The learning rate is adjusted by a cosine annealing schedule with warm restarts, where the initial period is set to 10 epochs and the minimum learning rate is $10^{-6}$. To ensure fairness, all TSM parameters, except for DLinear, are kept on the same order of magnitude (Table~\ref{tab:tsm}).

\begin{table}[htbp]
    \centering
    \caption{Hyperparameter configuration for the site-specific time series models. To ensure fairness in the comparison, we maintain the parameter size of the time series models at the same order of magnitude, except for DLinear.}
    \label{tab:tsm}
    \begin{tabular*}{\linewidth}{@{\extracolsep{\fill}}lccccc}
        \toprule
        Model & $e_{\text{layers}}$ & $d_{\text{model}}$ & $d_{\text{ff}}$ & $n_{\text{heads}}$ & Params (M) \\
        \midrule
        FEDformer \citep{zhou2022fedformer} & 5 & 256 & 512 & 8 & 3.421 \\
        DLinear \citep{zeng2023dlinear} & -- & -- & -- & -- & 0.327 \\
        NSTransformer \citep{liu2022nstransformer} & 5 & 256 & 512 & 8 & 3.909 \\
        TimesNet \citep{wu2022timesnet} & 3 & 32 & 64 & -- & 3.562 \\
        PatchTST \citep{nie2023patchtst} & 5 & 256 & 512 & 8 & 3.213 \\
        iTransformer \citep{liu2024itransformer} & 5 & 256 & 512 & 8 & 2.965 \\
        TimeXer \citep{wang2024timexer} & 5 & 256 & 512 & 8 & 4.421 \\
        TimeMixer \citep{wang2024timemixer} & 5 & 256 & 512 & -- & 4.873 \\
        \bottomrule
    \end{tabular*}
\end{table}

\subsubsection{Large time series models (LTSM)}\label{sec:ltsm_settings}

We evaluate 11 publicly available LTSM models in zero-shot mode on 127 unseen in-domain sites of the WTK dataset and 6 out-of-domain sites in the Kelmarsh dataset. These models were pretrained on large-scale, domain-independent time series corpora and applied directly without fine-tuning. The evaluated LTSM models show significant differences in architecture, covariate support, and output format:

\begin{itemize}[nosep]
\item \textbf{MOMENT-1-small / base / large} \citep{goswami2024moment}: encoder-based masked reconstruction models; used with univariate power input in this protocol; point output only.
\item \textbf{TimeMoE-50M / 200M} \citep{shi2024timemoe}: decoder-only MoE architectures; used with univariate power input in this protocol; point output only.
\item \textbf{Sundial-base-128M} \citep{zhang2025sundial}: flow/diffusion-style generative model with adaptive patching; used with univariate power input in this protocol; sample-path probabilistic output.
\item \textbf{TiRex} \citep{das2025tirex}: zero-shot forecasting model with enhanced in-context learning across long and short horizons; used with univariate power input in this protocol; quantile probabilistic output.
\item \textbf{TimesFM-2.5-200M} \citep{das2024timesfm}: decoder-only foundation model with patched time series tokenisation; used with univariate power input in this protocol; quantile probabilistic output.
\item \textbf{Chronos-2} \citep{ansari2025chronos2}: universal forecasting model supporting zero-shot multivariate and covariate-informed tasks; used with univariate power input in this protocol; quantile probabilistic output.
\item \textbf{Moirai-2.0-R-small} \citep{woo2024moirai}: universal forecasting Transformer supporting multivariate historical covariates; quantile probabilistic output.
\item \textbf{Timer-S1} \citep{liu2024timer}: billion-scale MoE Transformer with serial scaling; used with univariate power input in this protocol; quantile probabilistic output.
\end{itemize}

Among these, only Moirai-2.0-R-small natively supports multivariate historical covariates; all other LTSMs accept only univariate power input. Models that produce only point predictions (TimeMoE, MOMENT) participate in deterministic metric comparisons; those with quantile or sample-path outputs also participate in probabilistic comparisons. Future weather is disabled for all models to maintain protocol consistency.

\subsection{Experimental tasks}\label{sec:tasks}

\begin{figure}
  \centering
  \includegraphics[width=\textwidth]{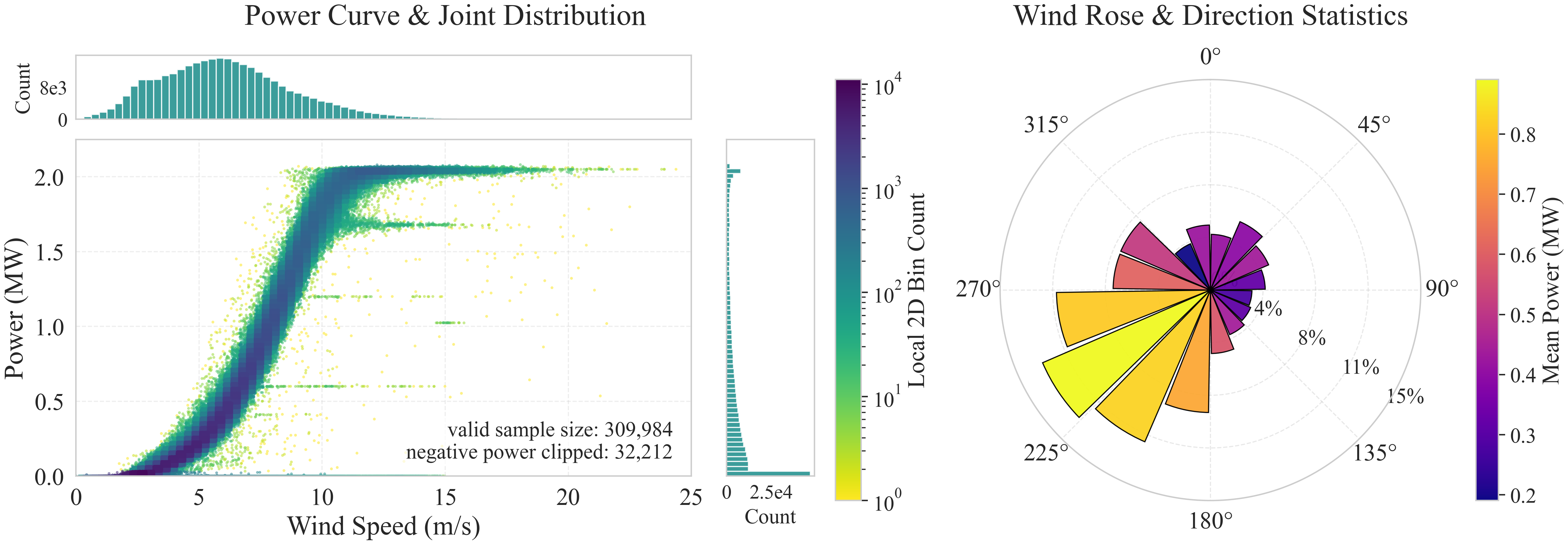}
  \caption{Visualization of power curve and wind rose at a representative site in the Kelmarsh dataset. $\mathbf{Left}$: joint distribution of wind speed and turbine-level power for turbine KWF1, with marginal histograms and a two-dimensional bin-count colour scale summarising 309,984 valid samples after clipping 32,212 negative-power records. $\mathbf{Right}$: wind rose summarising the directional frequency and wind-speed distribution for the same turbine.}
  \label{fig:KWF1_joint_distribution_and_wind_rose}
\end{figure}

Three groups of experiments are conducted to evaluate Tyan-WP from complementary perspectives:

\begin{enumerate}[label=(\arabic*)]
\item \textbf{WTK 10-site target-site supervised comparison.} 10 WTK sites are sampled randomly from the held-out pool of unseen in-domain sites using the 2011--2013 temporal partition: 2011 for training, 2012 for validation, and 2013 for testing. The meteorological variables of these 10 sites are independently z-standardized using single-site statistics computed from their own 2011--2012 records. Eight state-of-the-art TSM architectures are trained on each target site and evaluated on the 2013 test set, while Tyan-WP performs zero-shot inference on the same test set.

\item \textbf{WTK 127-site in-domain transfer.} All 127 held-out WTK sites serve as the in-domain zero-shot benchmark in the 2013 test year. To avoid information leakage, their meteorological variables are z-standardized using the global statistics of the WTK pretraining set. No model, including Tyan-WP and the public LTSMs, is trained or fine-tuned on these sites.

\item \textbf{Kelmarsh 6-site out-of-domain generalization.} The Kelmarsh wind farm \citep{plumley2025kelmarsh}, located in the United Kingdom, comprises six Senvion~MM92 turbines rated at 2.05\,MW. The raw 10-minute SCADA records span 2016--2024 and are resampled to 15-minute resolution. Meteorological variables are likewise z-standardized using the global statistics of the WTK pretraining set. Air density and surface pressure are synthesized from available variables through physical formulas before resampling. For the site embedding, Kelmarsh categorical attributes are not locally re-indexed; instead, terrain and ecoregion categories are mapped to WTK-compatible proxy classes (plain terrain and marine west coast forest), so that the categorical inputs remain aligned with the pretrained geography-ecology embedding vocabulary. Kelmarsh differs from WTK in country, turbine technology, rated capacity, data provenance, and native temporal resolution, providing a stringent out-of-domain transfer test. Figure~\ref{fig:KWF1_joint_distribution_and_wind_rose} illustrates the power-curve structure and wind-direction distribution of a representative Kelmarsh turbine.
\end{enumerate}

Table~\ref{tab:datasets} summarises the datasets used across these tasks.

\begin{table}[!t]
\caption{Summary of datasets used for pretraining and evaluation.}\label{tab:datasets}
\centering
\small
\begin{tabular*}{\textwidth}{@{\extracolsep{\fill}}lcccccc@{}}
\toprule
Dataset & Domain & Division method & Experimental subject \\
\midrule
WTK 126,565-site & In-domain & Pretrain: 2007--2012, Test: 2013 & Pretrained Tyan-WP \\
WTK 10-site & In-domain & Train: 2011, Validation: 2012, Test: 2013 & Site-specific supervised TSMs \\
WTK 127-site & In-domain & Test: 2013 & Zero-shot LTSMs \\
Kelmarsh 6-site & Out-of-domain & Test: 2024 & Zero-shot LTSMs \\
\bottomrule
\end{tabular*}
\end{table}

\subsection{Unified evaluation protocol}\label{sec:protocol}

All experiments share: input length $L{=}64$ (16\,h), prediction length $H{=}16$ (4\,h), 15-minute time steps, no future meteorological input, point prediction from $q_{0.5}$, and site-level metrics averaged with equal weight. Primary metrics are MAE, RMSE, $\mathrm{R^2}$, CRPS, and AQL, all computed in the MW power scale (except $\mathrm{R^2}$, which is dimensionless).

\subsubsection{Deterministic metrics}

Let $y_i$ denote the $i$-th ground-truth power value and $\hat{y}_i$ the corresponding point forecast, with $N$ total evaluation samples. Deterministic metrics include $\mathrm{MAE}$, $\mathrm{RMSE}$, and $\mathrm{R^2}$, and their calculation formulas are as follows:
\begin{equation}\label{eq:mae}
  \mathrm{MAE} = \frac{1}{N}\sum_{i=1}^{N}\bigl|y_i - \hat{y}_i\bigr|.
\end{equation}
\begin{equation}\label{eq:rmse}
  \mathrm{RMSE} = \sqrt{\frac{1}{N}\sum_{i=1}^{N}\bigl(y_i - \hat{y}_i\bigr)^2}.
\end{equation}
\begin{equation}\label{eq:r2}
  \mathrm{R^2} = 1 - \frac{\sum_{i=1}^{N}(y_i - \hat{y}_i)^2}{\sum_{i=1}^{N}(y_i - \bar{y})^2},
\end{equation}
where $\bar{y}=\frac{1}{N}\sum_{i=1}^{N}y_i$.

\subsubsection{Probabilistic metrics}

The pinball loss for quantile level~$q$ follows quantile regression \citep{koenker1978regression}:
\begin{equation}\label{eq:pinball}
  L_q(y,\hat{y}_q) = \max\!\bigl(q\,(y-\hat{y}_q),\;(q-1)(y-\hat{y}_q)\bigr).
\end{equation}

The average quantile loss (AQL) \citep{messner2020evaluation} is the mean pinball loss averaged over all $N$ samples, $H$ forecast horizons, and $|\mathcal{Q}|$ quantile levels:
\begin{equation}\label{eq:aql}
  \mathrm{AQL} = \frac{1}{N \cdot H \cdot |\mathcal{Q}|}\sum_{i=1}^{N}\sum_{h=1}^{H}\sum_{q\in\mathcal{Q}} L_q\!\bigl(y_{i,h},\;\hat{y}_{i,h,q}\bigr).
\end{equation}

Let $F_{i,h}(z)$ denote the predictive cumulative distribution function (CDF) for sample $i$ and forecast horizon $h$. The continuous ranked probability score (CRPS) \citep{gneiting2007strictly} measures the integrated squared distance between this predictive CDF and the empirical step CDF induced by the observed value $y_{i,h}$:
\begin{equation}\label{eq:crps}
  \mathrm{CRPS} =
  \frac{1}{N \cdot H}\sum_{i=1}^{N}\sum_{h=1}^{H}
  \int_{-\infty}^{\infty}
  \left(F_{i,h}(z)-\mathbf{1}\{y_{i,h}\le z\}\right)^2\,\mathrm{d}z .
\end{equation}
Thus, CRPS evaluates the full predictive distribution rather than a single quantile. For quantile- or sample-based model outputs, $F_{i,h}$ is constructed from the reported quantiles or empirical samples, and the integral is evaluated numerically on the MW power scale.

\section{Results and analysis}\label{sec:results}

\subsection{WTK 10-site: zero-shot Tyan-WP compared with target-site supervised TSMs}\label{sec:res_tsm}

\begin{table}[!htbp]
\caption{Comparison of deterministic and probabilistic forecasting results between WTK 10-site zero-shot Tyan-WP and site-specific supervised TSM baselines. Best in \textbf{bold} and second best underlined.}\label{tab:wtk10}
\centering
\small
\begin{tabular*}{\textwidth}{@{\extracolsep{\fill}}lccccc@{}}
\toprule
Model & MAE$\downarrow$ & RMSE$\downarrow$ & $\mathrm{R^2}\uparrow$ & CRPS$\downarrow$ & AQL$\downarrow$ \\
\midrule
FEDformer & 3.070 & 4.239 & 0.409 & 2.811 & 1.439 \\
DLinear & 2.051 & 3.172 & 0.666 & 1.652 & 0.873 \\
NSTransformer & 2.481 & 3.518 & 0.586 & 1.904 & 1.011 \\
TimesNet & 2.132 & 3.197 & 0.662 & 1.665 & 0.883 \\
PatchTST & 1.973 & 3.154 & 0.669 & \underline{1.496} & \underline{0.799} \\
iTransformer & \underline{1.951} & \underline{3.092} & \underline{0.683} & 1.500 & \underline{0.799} \\
TimeXer & 1.960 & 3.115 & 0.678 & 1.507 & 0.801 \\
TimeMixer & 1.980 & 3.161 & 0.668 & 1.580 & 0.837 \\
\textbf{Ours} & \textbf{1.692} & \textbf{2.810} & \textbf{0.737} & \textbf{1.227} & \textbf{0.661} \\
\bottomrule
\end{tabular*}
\end{table}

Table~\ref{tab:wtk10} reports the aggregate deterministic and probabilistic forecasting performance of zero-shot Tyan-WP against eight site-specific supervised TSM baselines on 10 held-out WTK sites in the 2013 test year. This comparison is stringent because the TSM baselines are trained with target-site historical data, whereas Tyan-WP is directly transferred to the target sites without supervised adaptation.

The table shows that Tyan-WP obtains the best value for all five metrics. Relative to the strongest site-specific supervised baselines for each metric, Tyan-WP reduces MAE by 13.3\%, RMSE by 9.1\%, CRPS by 18.0\%, and AQL by 17.3\%, while increasing $\mathrm{R^2}$ by 7.9\%. These results indicate that domain-specific pretraining can provide transferable information that is not fully recovered by site-specific training of conventional TSM architectures.

\begin{figure}
  \centering
  \includegraphics[width=\textwidth]{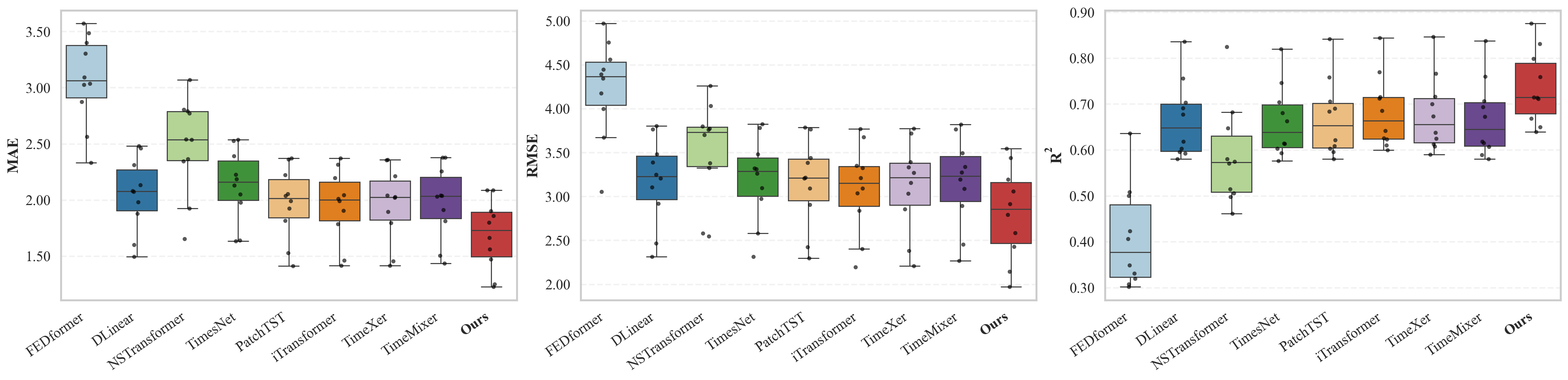}
  \captionof{figure}{A comparison of site-level deterministic distributions for zero-shot Tyan-WP and site-specific supervised TSM baselines at the WTK 10-site.}
  \label{fig:wtk10_deterministic_distributions}
\end{figure}

\begin{figure}
  \centering
  \includegraphics[width=\textwidth]{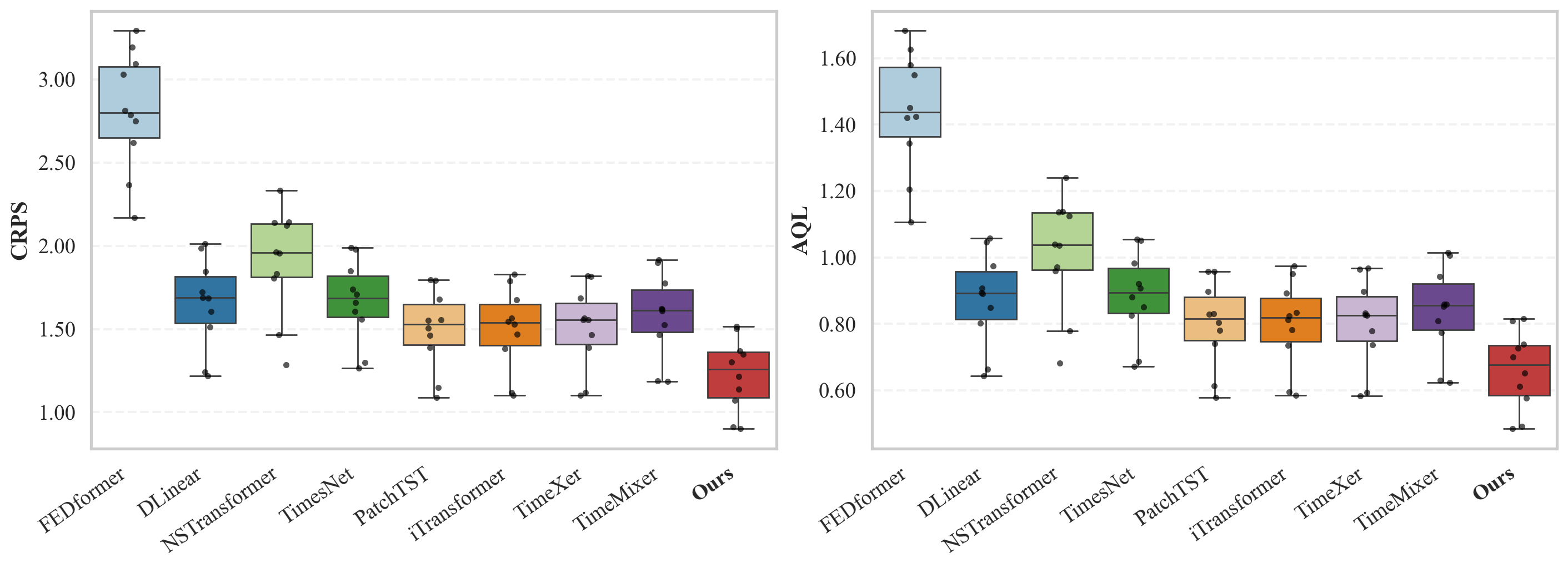}
  \captionof{figure}{A comparison of site-level probabilistic distributions for zero-shot Tyan-WP and site-specific supervised TSM baselines at the WTK 10-site.}
  \label{fig:wtk10_probabilistic_distributions}
\end{figure}

Figure~\ref{fig:wtk10_deterministic_distributions} complements the aggregate comparison by showing the site-level distributions of deterministic metrics. Tyan-WP is consistently shifted toward lower MAE and RMSE values, while the spread across sites confirms that the difficulty of the held-out locations is not uniform. Figure~\ref{fig:wtk10_probabilistic_distributions} shows a similar downward shift for CRPS and AQL, indicating that the probabilistic advantage is not caused by a single favourable site. Taken together, the table and figures support the conclusion that Tyan-WP improves both point accuracy and distributional calibration under zero-shot transfer.

\subsection{WTK 127-site in-domain transfer: Tyan-WP compared with generic LTSMs}\label{sec:res_ltsm}

\begin{table}[!htbp]
\caption{Comparison of deterministic and probabilistic forecasting results between WTK 127-site zero-shot Tyan-WP and generic LTSM baselines. Best in \textbf{bold} and second best underlined.}\label{tab:wtk127}
\centering
\footnotesize
\begin{tabular*}{\textwidth}{@{\extracolsep{\fill}}lccccc@{}}
\toprule
Model & MAE$\downarrow$ & RMSE$\downarrow$ & $\mathrm{R^2}\uparrow$ & CRPS$\downarrow$ & AQL$\downarrow$ \\
\midrule
Moment$_{\mathrm{small}}$ & 3.685 & 4.962 & 0.082 & -- & -- \\
Moment$_{\mathrm{base}}$ & 3.489 & 4.714 & 0.173 & -- & -- \\
Moment$_{\mathrm{large}}$ & 2.721 & 3.801 & 0.460 & -- & -- \\
TimeMoE$_{\mathrm{small}}$ & 2.204 & 3.298 & 0.590 & -- & -- \\
TimeMoE$_{\mathrm{base}}$ & 2.209 & 3.311 & 0.586 & -- & -- \\
Sundial & 2.142 & 3.300 & 0.588 & 1.786 & 0.925 \\
TiRex & \underline{1.916} & 3.189 & 0.612 & \underline{1.435} & \underline{0.768} \\
TimesFM 2.5 & 1.978 & 3.259 & 0.597 & 1.530 & 0.808 \\
Chronos 2 & 1.918 & 3.194 & 0.611 & 1.461 & 0.778 \\
Moirai 2.0 & 1.943 & \underline{3.123} & \underline{0.630} & 1.499 & 0.798 \\
Timer-S1 & 2.082 & 3.324 & 0.581 & 1.591 & 0.843 \\
\textbf{Ours} & \textbf{1.534} & \textbf{2.604} & \textbf{0.735} & \textbf{1.117} & \textbf{0.601} \\
\bottomrule
\end{tabular*}
\end{table}

Table~\ref{tab:wtk127} summarises the zero-shot comparison between Tyan-WP and generic LTSM baselines on 127 held-out WTK sites. Models that only produce point forecasts are included in the deterministic metrics, and their probabilistic entries are marked with ``--'' in the CRPS and AQL columns.

At this larger in-domain scale, Tyan-WP again achieves the best deterministic and probabilistic scores. Relative to the strongest generic LTSM baselines for each metric, Tyan-WP reduces MAE by 19.9\%, RMSE by 16.6\%, CRPS by 22.2\%, and AQL by 21.7\%, while increasing $\mathrm{R^2}$ by 16.7\%. Compared to LTSMs pretrained on extensive time series corpora, these improvements indicate that pretraining on domain-specific large-scale wind power sequence data, along with the two domain-specific module designs, static site embedding and power-aware meteorological fusion modeling, are crucial for enhancing the zero-shot wind power forecasting performance of newly installed turbines distributed across diverse terrains and meteorological conditions.

\begin{figure}
  \centering
  \includegraphics[width=\textwidth]{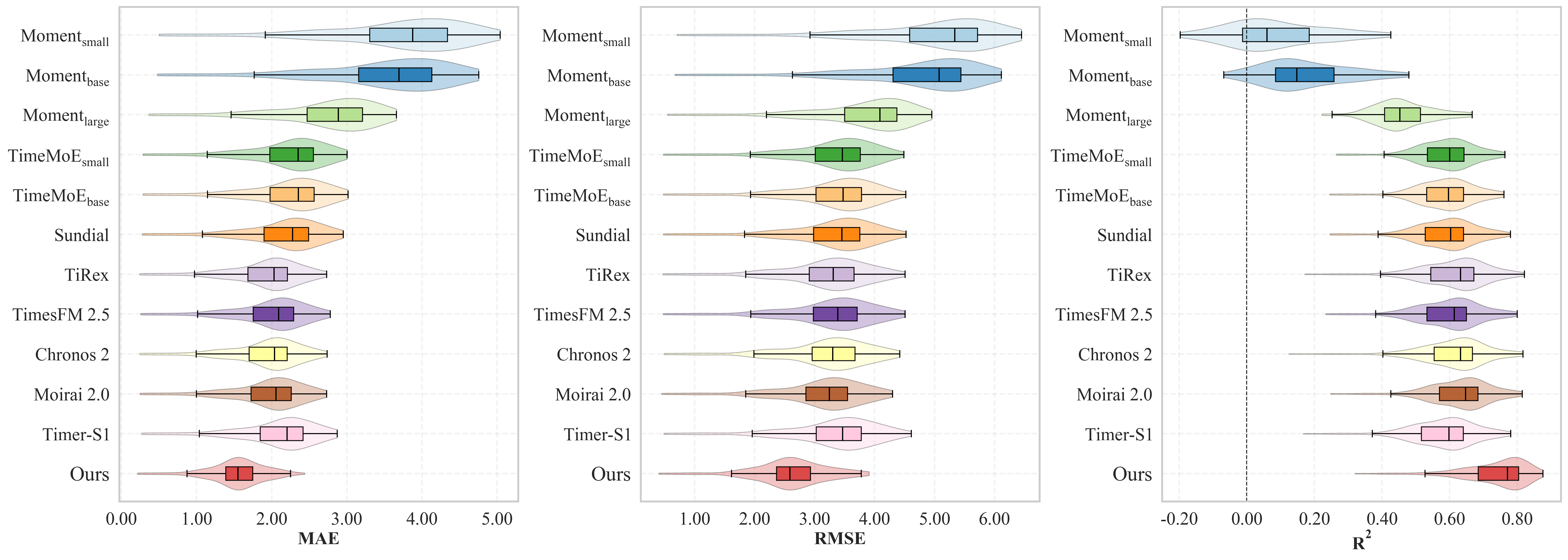}
  \captionof{figure}{A comparison of site-level deterministic distributions of zero-shot forecasting for Tyan-WP and generic LTSMs at the WTK 127-site.}
  \label{fig:wtk127_deterministic_distributions}
\end{figure}

\begin{figure}
  \centering
  \includegraphics[width=\textwidth]{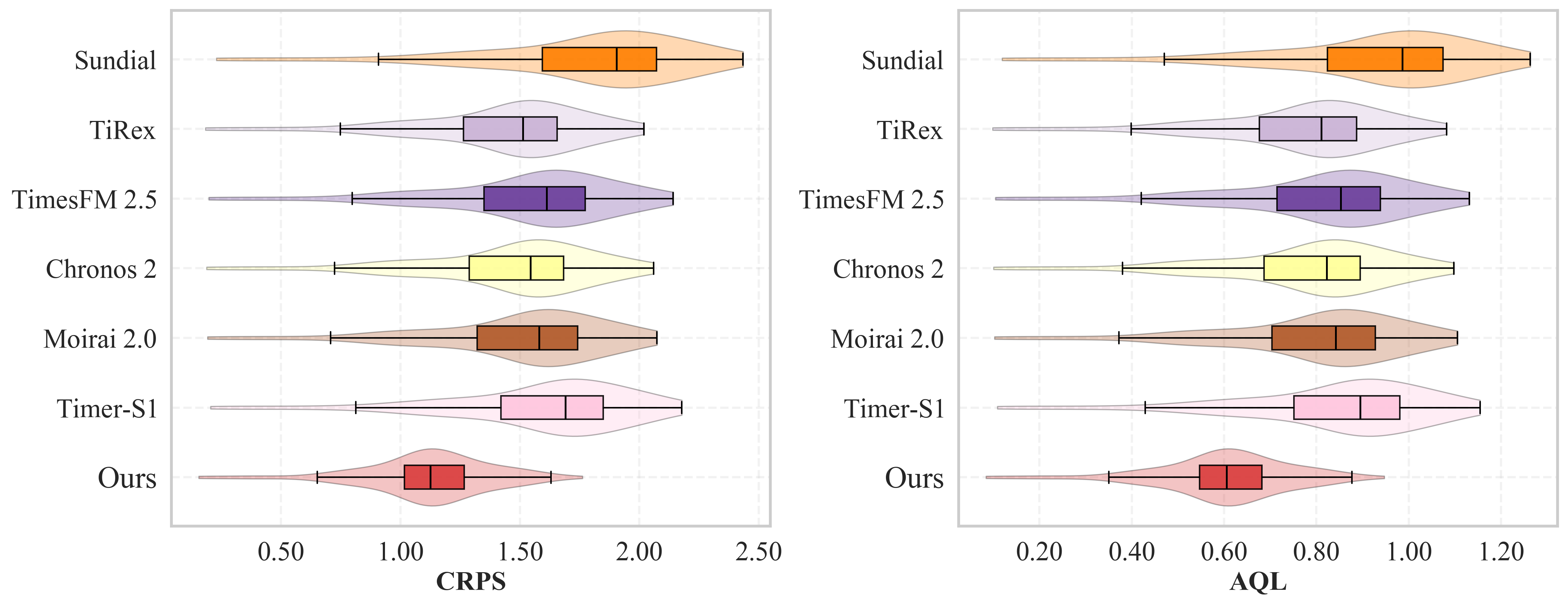}
  \captionof{figure}{A comparison of site-level probabilistic distributions of zero-shot forecasting for Tyan-WP and generic LTSMs at the WTK 127-site.}
  \label{fig:wtk127_probabilistic_distributions}
\end{figure}

Figure~\ref{fig:wtk127_deterministic_distributions} shows that Tyan-WP has the most favourable site-level distribution for deterministic forecasting, with lower MAE and RMSE distributions and a higher $\mathrm{R^2}$ distribution than the generic LTSM group. The improvement is visible across the distribution rather than only in the mean, which is important because the 127 sites cover diverse meteorological and power regimes. Figure~\ref{fig:wtk127_probabilistic_distributions} further shows that Tyan-WP shifts the CRPS and AQL distributions toward lower values compared with the probabilistic LTSM baselines. These distributional results are consistent with Table~\ref{tab:wtk127} and indicate robust in-domain zero-shot transfer across a broad WTK evaluation set.
\FloatBarrier

\subsection{Kelmarsh 6-site out-of-domain generalization: Tyan-WP compared with generic LTSMs}\label{sec:res_kelmarsh}

\begin{table}[!htbp]
\caption{Comparison of deterministic and probabilistic forecasting results between Kelmarsh 6-site zero-shot Tyan-WP and generic LTSM baselines. Best in \textbf{bold} and second best underlined.}\label{tab:kelmarsh}
\centering
\footnotesize
\begin{tabular*}{\textwidth}{@{\extracolsep{\fill}}lccccc@{}}
\toprule
Model & MAE$\downarrow$ & RMSE$\downarrow$ & $\mathrm{R^2}\uparrow$ & CRPS$\downarrow$ & AQL$\downarrow$ \\
\midrule
Moment$_{\mathrm{small}}$ & 0.346 & 0.495 & 0.328 & -- & -- \\
Moment$_{\mathrm{base}}$ & 0.325 & 0.467 & 0.401 & -- & -- \\
Moment$_{\mathrm{large}}$ & 0.262 & 0.383 & 0.597 & -- & -- \\
TimeMoE$_{\mathrm{small}}$ & 0.244 & 0.363 & 0.638 & -- & -- \\
TimeMoE$_{\mathrm{base}}$ & 0.245 & 0.365 & 0.635 & -- & -- \\
Sundial & 0.245 & 0.370 & 0.623 & 0.205 & 0.106 \\
TiRex & 0.221 & 0.345 & 0.673 & \underline{0.162} & \underline{0.087} \\
TimesFM 2.5 & 0.223 & 0.353 & 0.658 & 0.169 & 0.090 \\
Chronos 2 & \underline{0.218} & \underline{0.342} & 0.679 & \underline{0.162} & \underline{0.087} \\
Moirai 2.0 & \underline{0.218} & \underline{0.339} & \underline{0.685} & 0.164 & \underline{0.087} \\
Timer-S1 & 0.238 & 0.367 & 0.630 & 0.179 & 0.095 \\
\textbf{Ours} & \textbf{0.217} & \textbf{0.328} & \textbf{0.705} & \textbf{0.159} & \textbf{0.085} \\
\bottomrule
\end{tabular*}
\end{table}

\begin{figure}
  \centering
  \includegraphics[width=\textwidth]{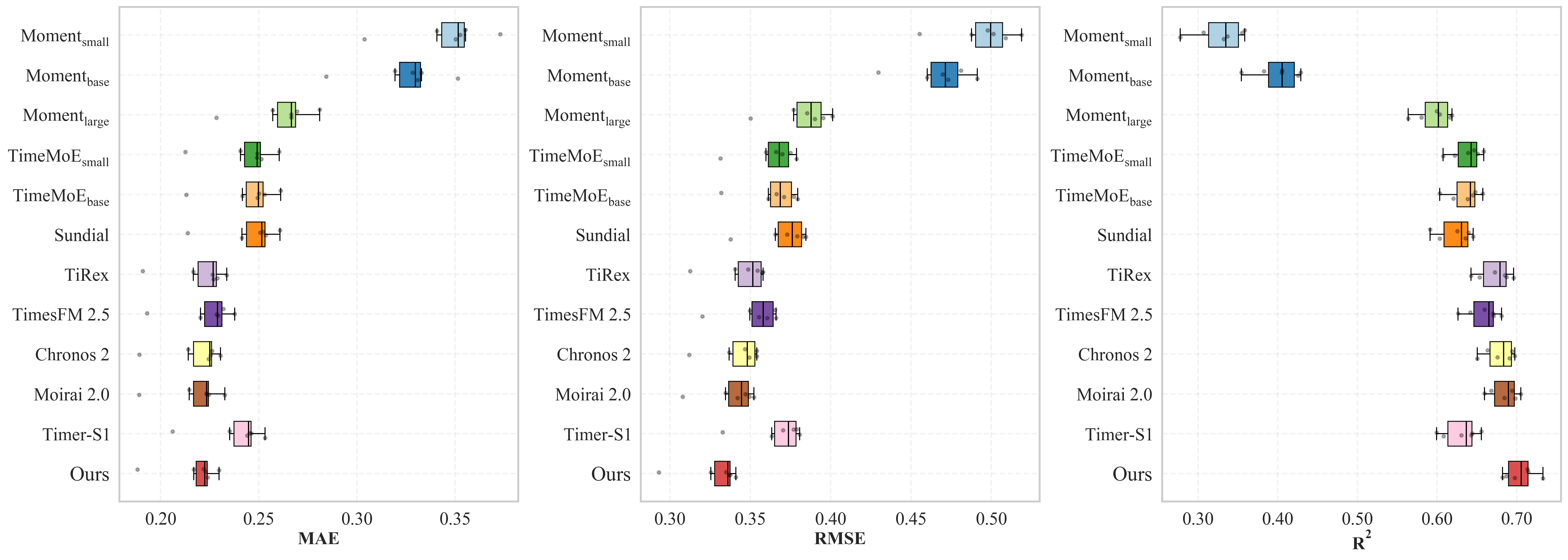}
  \captionof{figure}{A comparison of site-level deterministic distributions of zero-shot forecasting for Tyan-WP and generic LTSMs at the Kelmarsh 6-site.}
  \label{fig:kelmarsh_deterministic}
\end{figure}

\begin{figure}
  \centering
  \includegraphics[width=\textwidth]{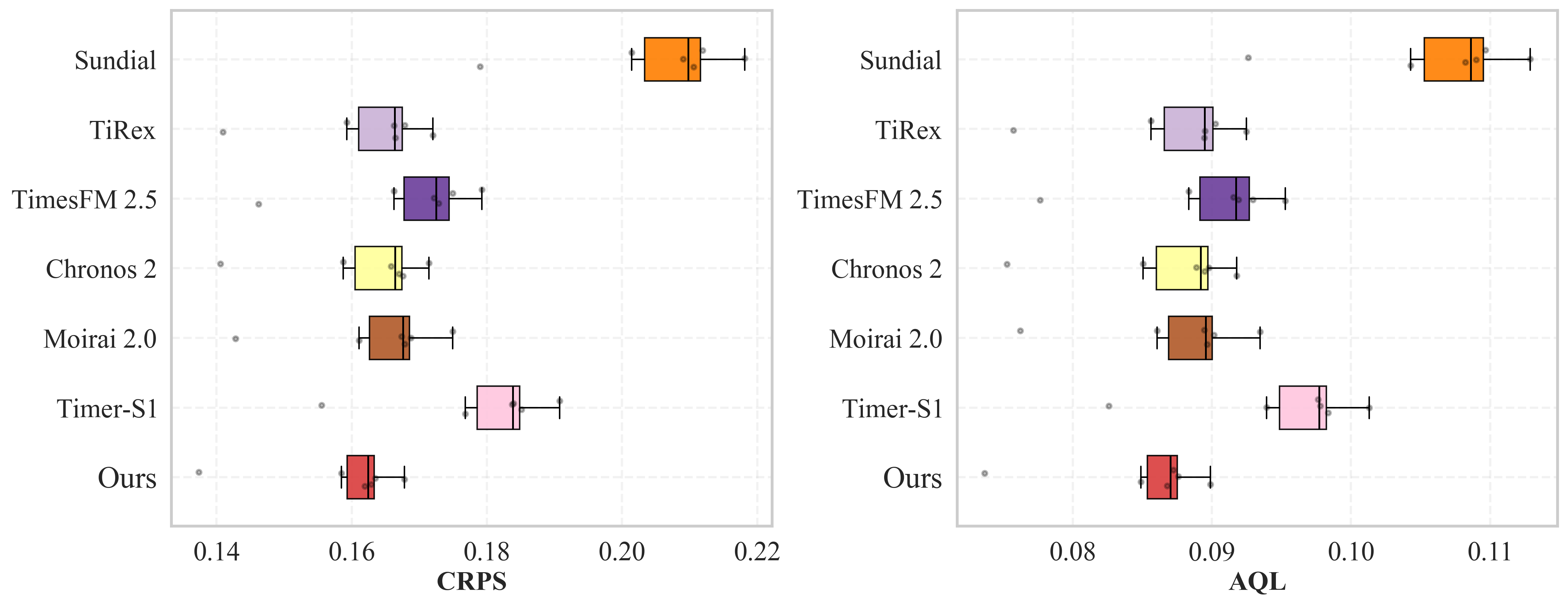}
  \captionof{figure}{A comparison of site-level probabilistic distributions of zero-shot forecasting for Tyan-WP and generic LTSMs at the Kelmarsh 6-site.}
  \label{fig:kelmarsh_probabilistic}
\end{figure}

Table~\ref{tab:kelmarsh} reports the out-of-domain zero-shot comparison on six Kelmarsh turbines. This setting is more challenging than the WTK evaluations because the target data differ in country, turbine rated capacity, and data source. More importantly, this benchmark involves a pronounced domain shift from the simulated U.S. WTK source domain to the real U.K. SCADA target domain, leading to substantial differences in data distribution and measurement characteristics.

Relative to the strongest generic LTSM baselines for each metric, Tyan-WP reduces MAE by approximately 0.5\%, RMSE by 3.2\%, CRPS by 1.9\%, and AQL by 2.3\%, while increasing $\mathrm{R^2}$ by approximately 2.9\%. These smaller improvements, compared with those observed on WTK, likely reflect the larger distribution shift between the WTK pretraining domain and the Kelmarsh target domain, which weakens the direct transferability of pretrained wind-power representations across domains.

Figure~\ref{fig:kelmarsh_deterministic} shows that several generic LTSMs become competitive on deterministic metrics under the Kelmarsh transfer setting, especially TiRex, Chronos 2, and Moirai 2.0. Tyan-WP remains within the best-performing cluster and provides the lowest aggregate RMSE together with the highest aggregate $\mathrm{R^2}$. Figure~\ref{fig:kelmarsh_probabilistic} indicates that Tyan-WP also has compact CRPS and AQL distributions near the lower end of the model group. These observations support the aggregate results in Table~\ref{tab:kelmarsh} and suggest that the proposed model retains useful domain-specific information when transferred from simulated U.S. WTK data to real U.K. SCADA data.
\FloatBarrier

\subsection{Ablation study}\label{sec:ablation}

\begin{table}[!htbp]
\caption{Ablation comparison of Tyan-WP variants on WTK 127-site and Kelmarsh 6-site zero-shot benchmarks. Values in parentheses denote relative changes from Ours. Best values within each dataset are in \textbf{bold}.}\label{tab:ablation}
\centering
\footnotesize
\setlength{\tabcolsep}{1.0pt}
\renewcommand{\arraystretch}{1.02}
\begin{tabular*}{\textwidth}{@{\extracolsep{\fill}}p{0.11\textwidth}p{0.24\textwidth}ccccc@{}}
\toprule
Dataset & Variant & MAE$\downarrow$ & RMSE$\downarrow$ & $\mathrm{R^2}\uparrow$ & CRPS$\downarrow$ & AQL$\downarrow$ \\
\midrule
\multirow{5}{0.11\textwidth}{WTK 127-site}
& Intra-channel attention module & 1.588~(+3.5\%) & 2.668~(+2.5\%) & 0.722~(-1.8\%) & 1.156~(+3.5\%) & 0.621~(+3.3\%) \\
& Channel-independent module & 1.567~(+2.2\%) & 2.653~(+1.9\%) & 0.723~(-1.6\%) & 1.140~(+2.1\%) & 0.613~(+2.0\%) \\
& Without site embedding & 1.561~(+1.8\%) & 2.634~(+1.2\%) & 0.729~(-0.8\%) & 1.135~(+1.6\%) & 0.611~(+1.7\%) \\
& Dense FFN encoder & 1.543~(+0.6\%) & 2.620~(+0.6\%) & 0.732~(-0.4\%) & 1.124~(+0.6\%) & 0.604~(+0.5\%) \\
& \textbf{Ours} & \textbf{1.534} & \textbf{2.604} & \textbf{0.735} & \textbf{1.117} & \textbf{0.601} \\
\midrule
\multirow{5}{0.11\textwidth}{Kelmarsh 6-site}
& Intra-channel attention module & 0.226~(+4.1\%) & 0.342~(+4.3\%) & 0.680~(-3.5\%) & 0.164~(+3.1\%) & 0.088~(+3.5\%) \\
& Channel-independent module & 0.229~(+5.5\%) & 0.349~(+6.4\%) & 0.666~(-5.5\%) & 0.167~(+5.0\%) & 0.089~(+4.7\%) \\
& Without site embedding & 0.219~(+0.9\%) & 0.335~(+2.1\%) & 0.692~(-1.8\%) & 0.160~(+0.6\%) & 0.086~(+1.2\%) \\
& Dense FFN encoder & 0.263~(+21.2\%) & 0.378~(+15.2\%) & 0.608~(-13.8\%) & 0.191~(+20.1\%) & 0.103~(+21.2\%) \\
& \textbf{Ours} & \textbf{0.217} & \textbf{0.328} & \textbf{0.705} & \textbf{0.159} & \textbf{0.085} \\
\bottomrule
\end{tabular*}
\end{table}

To individually evaluate the contribution of each architectural component, we evaluate four structural ablation schemes on the in-domain WTK 127-site and out-of-domain Kelmarsh 6-site zero-shot benchmarks. These variants sequentially replace the power and meteorological features from the designed channel fusion module with intra-channel attention or channel-independent modules, remove site embeddings, and replace the sparse mixture of experts (MoE) in the encoder with a dense feedforward network (FFN). The full model results (Ours) in Table~\ref{tab:ablation} are taken from Tables~\ref{tab:wtk127} and~\ref{tab:kelmarsh}.

Table~\ref{tab:ablation} shows that the full model obtains the best results on both datasets, indicating that the power-aware meteorological fusion (PAMF) module, site-level geography-ecology embedding, and sparse expert encoder are all useful for zero-shot wind power forecasting. The magnitude of their contributions differs across datasets, reflecting different structural requirements for in-domain transfer and out-of-domain generalization.

The PAMF module provides stable gains on both datasets. On WTK 127-site, replacing it with the intra-channel attention module increases MAE, RMSE, CRPS, and AQL by approximately 3.5\%, 2.5\%, 3.5\%, and 3.3\%, respectively, while reducing $\mathrm{R^2}$ by approximately 1.8\%. Replacing it with the channel-independent module increases these error metrics by approximately 2.2\%, 1.9\%, 2.1\%, and 2.0\%, with an approximately 1.6\% decrease in $\mathrm{R^2}$. On Kelmarsh, the effect is stronger: the intra-channel attention module increases MAE, RMSE, CRPS, and AQL by approximately 4.1\%, 4.3\%, 3.1\%, and 3.5\%, and reduces $\mathrm{R^2}$ by approximately 3.5\%; the channel-independent module further increases these metrics by approximately 5.5\%, 6.4\%, 5.0\%, and 4.7\%, with an approximately 5.5\% decrease in $\mathrm{R^2}$. These results indicate that neither intra-channel attention nor channel-independent processing can fully replace power-aware meteorological fusion. The PAMF module first reinforces short-term power dynamics and then integrates them with complementary meteorological information through power-conditioned attention and gating, which helps the model emphasize weather changes that are informative for future power variation. This mechanism is more important on Kelmarsh because all WTK variables are produced by numerical simulation or calculation-based synthesis, whereas Kelmarsh comes from real SCADA records. In other words, compared with synthetic data, real data contain richer complementary meteorological information, which is more beneficial for predicting future wind power.

The contribution of site embedding mainly appears in cross-site heterogeneity modelling. On WTK 127-site, removing site embedding increases MAE, RMSE, CRPS, and AQL by approximately 1.8\%, 1.2\%, 1.6\%, and 1.7\%, respectively, and reduces $\mathrm{R^2}$ by approximately 0.8\%. Because the 127 WTK sites cover broad geographical areas and wind-regime conditions, geographical location, terrain, and ecoregion priors help distinguish site-specific wind-resource backgrounds and power responses. On Kelmarsh, removing site embedding increases MAE, RMSE, CRPS, and AQL by approximately 0.9\%, 2.1\%, 0.6\%, and 1.2\%, respectively, and reduces $\mathrm{R^2}$ by approximately 1.8\%. The smaller contribution is mainly because Kelmarsh contains only six turbines from the same wind farm, with limited spatial and terrain diversity, and its terrain and ecoregion categories use WTK-compatible proxy embeddings, making the static site prior less discriminative than in WTK 127-site.

The sparse expert encoder contributes most strongly to out-of-domain generalization. On WTK 127-site, replacing the sparse MoE encoder with a dense FFN encoder increases MAE, RMSE, CRPS, and AQL by only approximately 0.6\%, 0.6\%, 0.6\%, and 0.5\%, respectively, and reduces $\mathrm{R^2}$ by approximately 0.4\%, suggesting that a shared dense FFN already covers most common wind-regime patterns under same-domain WTK testing. On Kelmarsh, the same replacement increases MAE, RMSE, CRPS, and AQL by approximately 21.2\%, 15.2\%, 20.1\%, and 21.2\%, respectively, and reduces $\mathrm{R^2}$ by approximately 13.8\%. Compared with a single dense FFN, MoE can route wind-regime segments to different experts according to token representations, which better handles different wind-regime patterns such as rapid ramps, low-wind conditions, and high-wind conditions. Compared with WTK in-domain transfer, Kelmarsh out-of-domain generalization across region, data source, and turbine type relies more strongly on this wind-regime-adaptive representation capacity.

Overall, the PAMF module is the most consistently effective structure across the two tasks because it strengthens the interaction modelling between recent power dynamics and complementary meteorological information. The benefit of site embedding increases with spatial diversity among target sites. The sparse expert encoder contributes most under out-of-domain generalization, showing that sparse expert routing improves the model's adaptability to wind-regime heterogeneity and data-source shift. The deterministic and probabilistic metrics show consistent trends across ablations, indicating that these modules improve not only median point forecasting but also the overall quality of the predictive distribution.

\section{Conclusion and future work}\label{sec:conclusion}

This paper presents Tyan-WP, the first wind power foundation model for ultra-short-term probabilistic forecasting. In addition to pretraining on a large-scale WTK dataset, Tyan-WP improves zero-shot forecasting through static site embedding and a power-aware meteorological fusion (PAMF) module. The static site embedding encodes latitude, longitude, terrain, and ecoregion metadata, while the PAMF module models dependencies between historical power and meteorological covariates. The fused representations are then processed by a sparse expert encoder and a horizon-quantile forecasting head to produce direct multi-horizon probabilistic forecasts without target-site supervised adaptation.

Under the evaluated in-domain and out-of-domain zero-shot protocols, Tyan-WP delivers the best overall performance across the compared site-specific TSMs and generic LTSMs, ranking first on all deterministic and probabilistic metrics in every setting. The ablation results further verify the contribution of the proposed wind-specific structures. The PAMF module provides the most consistent gains across in-domain transfer and out-of-domain generalization, site embedding improves cross-site heterogeneity modelling, and the sparse expert encoder is particularly important for out-of-domain Kelmarsh generalization.

These results demonstrate that domain-specific pretraining with wind-specific structural priors offers a practical pathway for rapid turbine onboarding. Newly commissioned, sensor-equipped wind turbines can receive competitive probabilistic forecasts immediately without site-specific supervised training, supporting uncertainty-aware real-time dispatch and operational risk management. Looking ahead, we plan to develop a foundational model for short-to-medium-term wind power forecasting that integrates numerical weather predictions or forecasts from large weather forecast models such as Pangu-Weather \citep{bi2023panguweather} and FengWu \citep{chen2025fengwu}, through innovative network architecture design.


\section*{Declaration of competing interest}

The authors declare that they have no known competing financial interests or personal relationships that could have appeared to influence the work reported in this paper.

\section*{Data availability}

The WTK dataset is publicly available at https://doi.org/10.1016/j.apenergy.2015.03.121. The Kelmarsh dataset is publicly available at https://doi.org/10.5281/zenodo.16807551.

\section*{Code availability}

The source code supporting this study is publicly available at https://github.com/USTC-AI4EEE/Tyan-WP, and the model weights are publicly available at https://huggingface.co/USTC-CILab/Tyan-WP.

\section*{Acknowledgment}

This research is supported by the National Science and Technology Major Project of China (Grant No. 2025ZD0805500).




\bibliographystyle{windpowerfm-vancouver}
\bibliography{windpowerfm-refs}

\end{document}